\documentclass[runningheads]{llncs}
\usepackage[T1]{fontenc}
\usepackage{tabularx}
\usepackage{amssymb}
\usepackage{amsmath,amsfonts}
\usepackage{graphicx}
\usepackage{xcolor}
\usepackage{comment}
\definecolor{newcolor}{rgb}{.8,.349,.1}

\begin{document}
\title{What can we learn about a generated image corrupting its latent representation?}
%
%
\author{Agnieszka Tomczak\inst{1,2} \and 
Aarushi Gupta\inst{3} \and 
Slobodan Ilic\inst{1,2} \and 
Nassir Navab\inst{1,4} \and 
Shadi Albarqouni\inst{4,5,1}\orcidID{0000-0003-2157-2211}} 
\authorrunning{A. Tomczak et al.}
%
\institute{Faculty of Informatics, Technical University of Munich, Munich, Germany \and
Siemens AG, Munich, Germany\\ \and
Indian Institute of Technology Delhi, New Delhi, India\\ \and
Clinic for Diagnostic and Interventional Radiology, University Hospital Bonn, Germany \\ \and
Helmholtz AI, Helmholz Zentrum Munich, Munich, Germany
\\
Corresponding author: Agnieszka Tomczak: a.tomczak@tum.de
}
\maketitle              
\begin{abstract}
Generative adversarial networks (GANs) offer an effective solution to the image-to-image translation problem, thereby allowing for new possibilities in medical imaging. They can translate images from one imaging modality to another at a low cost. For unpaired datasets, they rely mostly on cycle loss. Despite its effectiveness in learning the underlying data distribution, it can lead to a discrepancy between input and output data. The purpose of this work is to investigate the hypothesis that we can predict image quality based on its latent representation in the GANs bottleneck. We achieve this by corrupting the latent representation with noise and generating multiple outputs. The degree of differences between them is interpreted as the strength of the representation: the more robust the latent representation, the fewer changes in the output image the corruption causes. Our results demonstrate that our proposed method has the ability to i) predict uncertain parts of synthesized images, and ii) identify samples that may not be reliable for downstream tasks, e.g., liver segmentation task.

\end{abstract}
\section{Introduction}

Generative Adversarial Networks (GANs)~\cite{gan} are state-of-the-art methods for image-to-image translation problems. It has been found that GANs are a promising technique for generating images of one modality based on another. Creating such images in a clinical setting could be highly effective, but only if the images retain anatomical details and serve the downstream task. As Cohen et al. ~\cite{cohen2018distribution} described, GANs based on cycle consistency can "hallucinate" features (for example tumors) causing potentially wrong diagnoses.
A clinically useful example of modality translation would be generating Computer Tomography (CT) images from Magnetic Resonance Imaging (MRI) scans, and vice versa. This problem has already been investigated multiple times~\cite{chen2021targan,zhang2018translating,emami2021sagan,Chen2018TechnicalNU,ge2019unpaired,yang2019unsupervised} as an image-to-image translation task with multiple approaches: some of them focusing on shape consistency to preserve anatomical features~\cite{yu2019EaGAN,zhang2018translating,emami2021sagan,ge2019unpaired,horvath2022metgan}, others proposing a multimodal approach to deal with the scalability issues~\cite{Xin2020Multi,huang2019cocagan,ShenZWXPTHSMCWX21}. 
Nevertheless, the main challenge remains: how to determine when a generative adversarial network can be trusted? The issue has a considerable impact on the medical field, where generated images with fabricated features have no clinical value. Recently, Upadhyay et al.~\cite{upadhyay2021uncerguidedi2i} tackled this problem by predicting not only the output images but also the corresponding aleatoric uncertainty and then using it to guide the GAN to improve the final output. Their method requires changes in the optimization process (additional loss terms) and network architecture (additional output). They showed that for paired datasets it results in improved image quality, but did not clearly address the point that in medical imaging the visual quality of images does not always transfer to the performance on a downstream task.
Our goal was to examine this problem from a different perspective and test the hypothesis: the more robust the image representation, the better the quality of the generated output and the end result. To this end, we present a noise injection technique that allows generating multiple outputs, thus quantifying the differences between these outputs and providing a confidence score that can be used to determine the uncertain parts of the generated image, the quality of the generated sample, and to some extent their impact on a downstream task.
\\
\section{Methodology}
We design a method to test the assumption that the stronger the latent representation the better the quality of a generated image. In order to check the validity of this statement, we corrupt the latent representation of an image with noise drawn from normal distribution and see how it influences the generated output image. In other terms, given an image $x\in X$, domain $y \in Y$ and a Generative Adversarial Network $G$, we assume a hidden representation $h= E(x)$ with dimensions $n,m,l$, where $E$ stands for the encoding part of $G$, and $D$ for the decoding part. We denote the generated image as $\hat{x}$. Next, we construct $k$ corrupted representation latent codes $\hat{h}$, adding to $h$ noise vector $\eta$:
\begin{equation}
\begin{gathered}
\eta_{1,..,k} \thicksim \mathcal{N}(0,\alpha \sigma_{h_{1,...,l}}^{2})
\end{gathered}
\end{equation}
where  $\sigma_{h_{1,...,l}}^2$ is channel-wise standard deviation of input representation $h$. We can control the noise level with factor $\alpha$. Before we add the noise vector, we eliminate the background noise with operation $bin$ by masking it with zeros for all the channels where the output pixels are equal to zero, so do not contain any information.
\begin{equation}
\begin{gathered}
bin(h) = h[\hat{x}>0] \\
\hat{h}_{1,...,k} = h_{1}+bin(h_{1}) \eta_{1},...,h_{k}+bin(h_{k}) \eta_{k}
\end{gathered}
\end{equation}
Having now multiple representations for a single input image we can pass them to decoder $D$ and generate multiple outputs:
\begin{equation}
\begin{gathered}
\hat{x}_{1,...,k} = D(\hat{h}_{1}),...,D(\hat{h}_{k})
\end{gathered}
\end{equation}
We use the multiple outputs to quantify the uncertainty connected with the representation of a given image. We calculate two scores: the variance (the average of the squared deviations from the mean) $\gamma$ of our $k$ generated images
\begin{equation}\label{eq:var}
\begin{gathered}
\gamma = Var(\hat{x_1},..,\hat{x_k})
\end{gathered}
\end{equation}
and the Mutual Information (MI) between the multiple outputs and our primary output $\hat{x}$ produced without noise injection.
\begin{equation}\label{eq:mi}
\begin{gathered}
MI(X;Y) = \sum_{y\in Y}\sum_{x \in X}p(x,y)\log\Big(\frac{p(x,y)}{p(x)p(y)}\Big) \\
\delta = \frac{1}{k}\sum_0^k MI\big(\hat{x},(\hat{x_i})\big)
\end{gathered}
\end{equation}
We interpret the $\gamma$ and $\delta$ as the measures of the representation quality. The variance $\gamma$ can be considered as an uncertainty score - the higher the variance of generated outputs with the corrupt representations, the more uncertain the encoder is about produced representation. On the other hand, the MI $\delta$ score can be interpreted as a confidence score, quantifying how much information is preserved between the original output $\hat{x}$ and the outputs produced from corrupted representations $\hat{x_1},..., \hat{x_k}$. We calculate the MI based on a joint (2D) histogram, with number of bins equal to $\lfloor \sqrt{n/5} \rfloor $, where n is a number of pixels per image as proposed by~\cite{cellucci2005bins}.

\section{Experiments and Results}
We conducted a number of experiments using state-of-the-art architectures with the goal of demonstrating the effectiveness of our proposed method and confirming our hypothesis that the stronger the latent representation, the better and more reliable the image quality. Our proposed method was evaluated on two publicly available datasets, namely CHAOS~\cite{CHAOS2021} and LiTS~\cite{lits} datasets.

\subsection{Network Architectures and implementation details}

\paragraph{TarGAN}
As our main baseline we use TarGAN~\cite{chen2021targan} network which uses a shape-consistency loss to preserve the shape of the liver. We trained the model for 100 epochs. We kept all the parameters unchanged with respect to the official implementation provided by the authors of TarGAN. We use PyTorch 1.10 to implement all the models and experiments.
During inference, we constructed $k=10$ corrupted representation with noise level $\alpha=3$ and used them for evaluation of our method.
\paragraph{UP-GAN}
We adapted the UP-GAN network from \cite{upadhyay2021uncerguidedi2i} to run on unpaired dataset as shown in \cite{upadhyay2021robustness}.
UP-GAN uses an uncertainty guided loss along the standard cycle loss during training. The uncertainty loss defined for UP-GAN was used in every component of cycleGAN - identity loss and cycle loss for training of both generators. We kept the learning rate at $10^{-4}$ for T1 to T2 transfer and at $10^{-3}$ for CT to T1 and T2 transfer. We tuned the hyperparameters in the following manner: 0.5 for each of the discriminator losses, while the generators had a factor 1 with their cycle losses, 0.01 with the uncertainty cycle loss and factors of 0.1 and 0.0015 with identity losses. We trained all the three models for 100 epochs.

\paragraph{Datasets.}
We use data of each modality (CT, T1 and T2) from 20 different patients provided by the publicly available CHAOS19 dataset~\cite{CHAOS2021}. We randomly selected 50\% of the dataset to be the training set and used the rest for testing. We followed \cite{chen2021targan} in setting liver as the target area, as the CT scans only have liver annotations.
%
%
Besides, we used LiTS~\cite{lits} dataset to evaluate our method on the pathological samples. The dataset contains CT scans of patients with liver tumors and corresponding segmentation masks. All images were resized to the size of $256\times256$.

\paragraph{Evaluation metrics.}
We use FID~\cite{fid} score to evaluate the quality of generated images and DICE score to evaluate segmentation results.  We calculate the FID score using features extracted with Model Genesis~\cite{zhou2019models}. 




\subsection{Can we use the noise injections to identify uncertain parts of a synthesized image?}

\begin{figure*}[h]
\centering
\includegraphics[width=\textwidth]{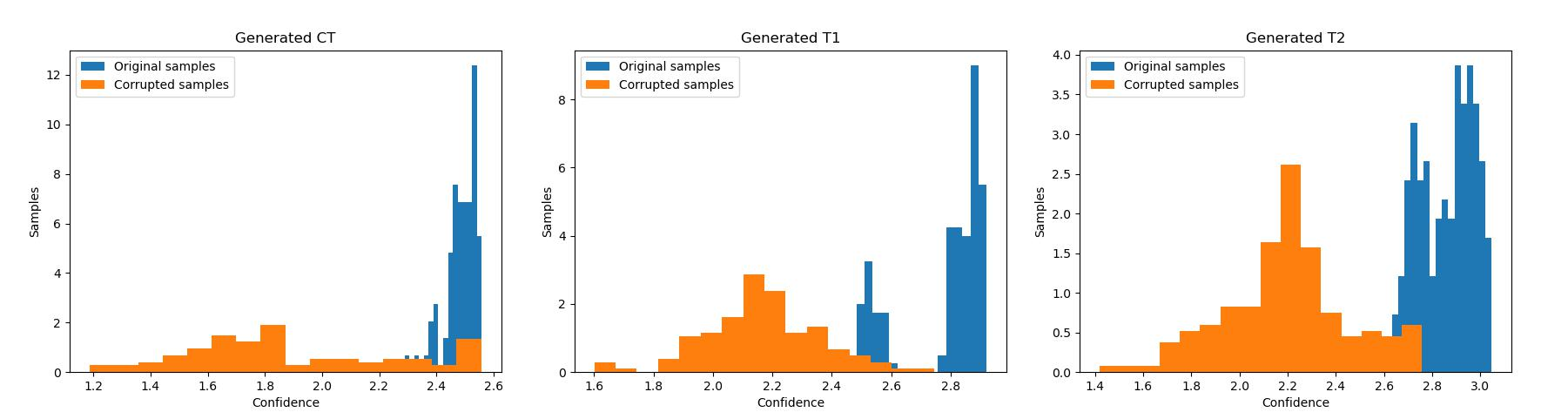} 
\caption{Histograms of confidence values for original images and images corrupted by blacking out 50x50 pixel square. We can see significant drop of confidence for the corrupted inputs in all of the target modalities.}
\label{fig:histograms}
\end{figure*}

\begin{figure*}[t]
\centering
\includegraphics[scale=0.29]{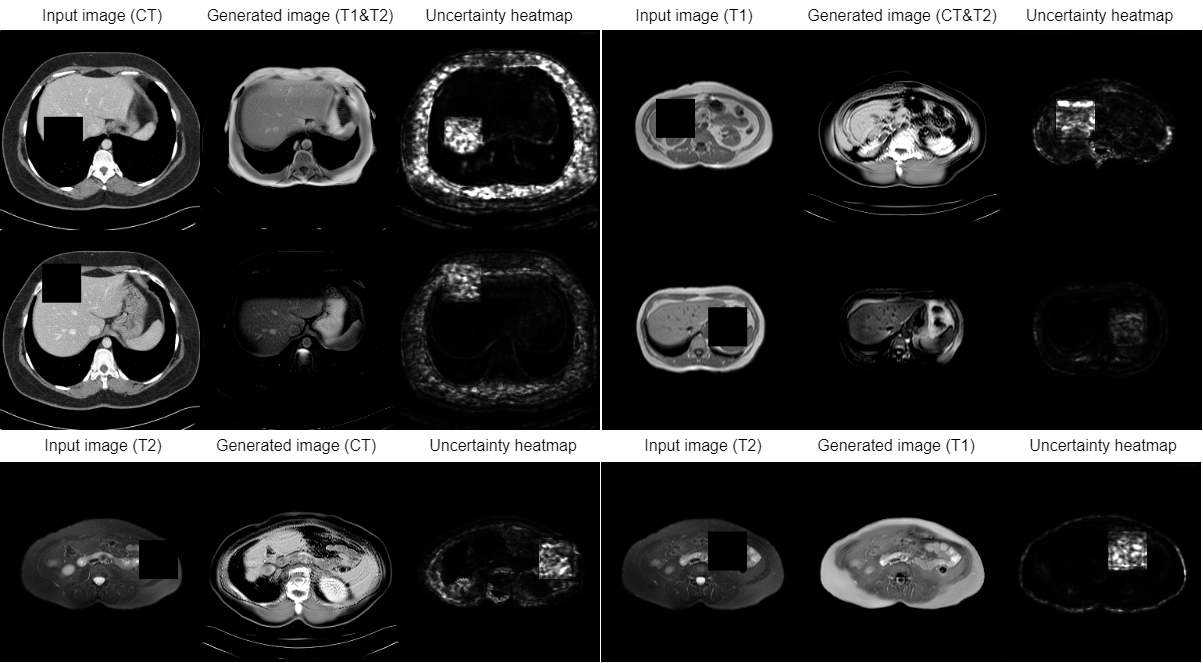} 
\caption{Samples generated using input with randomly blacked out $50\times50$ pixel patch. We can see on uncertainty heat maps the lack of confidence when generating this part of the images. It is not reflected in the output image.}
\label{fig:anomalies}
\end{figure*}

First, we conduct a sanity check experiment by blacking out a random $50\times50$ pixel patch from the input images (perturbed input) and measuring the proposed uncertainty and confidence scores on the corresponding synthesized images. Table~\ref{tab:anomalies} report the mean, median and variance of both uncertainty score ($\gamma$ in eq.\ref{eq:var}) and confidence score ($\delta$ in eq.\ref{eq:mi}) on both the original and perturbed images. One could observe that perturbed input has large variance and low confidence compared to the original input. This has been nicely visualized in Figure~\ref{fig:histograms} where the confidence scores of the perturbed corrupted images are much lower than the corresponding ones for the original images. This demonstrates the effectiveness of our proposed method in detecting uncertain synthesized images, e.g., perturbed images. The results suggest the possibility of finding a potential confidence threshold to eliminate uncertain synthesized images. Surprisingly, the model was able to nicely synthesize perturbed images hallucinating and replacing the masked regions with reasonable healthy tissues. Nevertheless, our heatmaps were able to capture such uncertainty as shown in Figure~\ref{fig:anomalies}. 

\begin{figure*}[t]
\centering
\includegraphics[width=\textwidth]{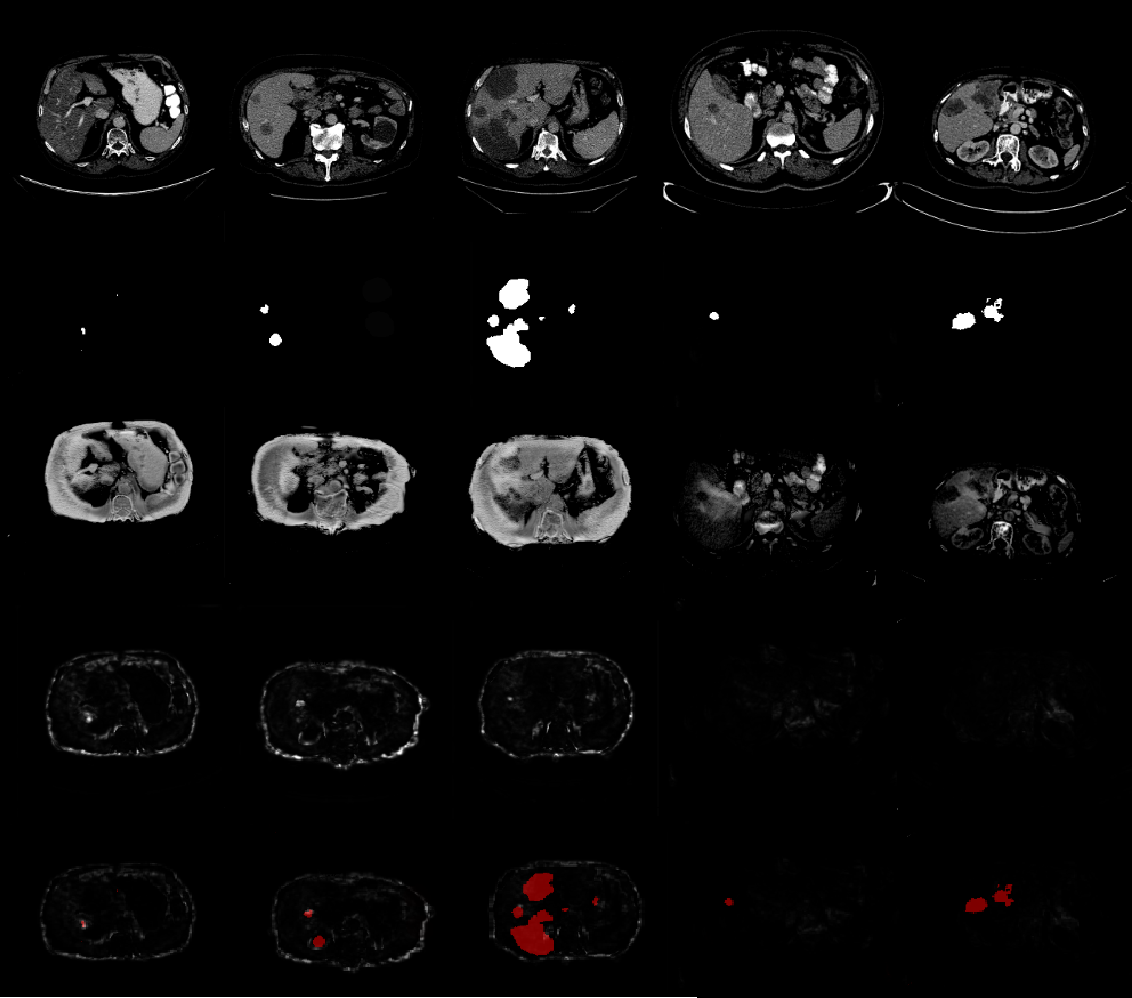} 
\caption{CT slides from LiTS dataset containing tumor pathologies. From top to bottom, we show the input CT slice, corresponding tumor segmentation map, generated T1 or T2 images, the uncertainty heat map and the overlay of segmentation mask and uncertainty map. }
\label{fig:lits}
\end{figure*}

To validate our proposed method in more realistic clinical setting, we run the inference on the LiTS dataset, which consists of CT scans with tumors present. While one would expect to see high uncertainty for input images with tumors, since the model was trained on healthy data, we only observed this for the translation from CT to T1 and only for small tumors (\emph{cf.} first two columns in Figure~\ref{fig:lits}). This was not the case for bigger tumors (third column) and for CT to T2 translation (last two columns). It seems that the network was confidently preserving big tumors with T1 target modality and all of the pathologies with T2 target modality, while small lesions in translation from CT to T1 - the ones that were not generated and probably filtered out as artifacts - caused spike in the uncertainty value.


\begin{table}[t]
\caption{The variance of multiple outputs with corrupted representation increases as we introduce a black patch on the input.}
\begin{tabular}{p{0.33\textwidth}p{0.33\textwidth}p{0.33\textwidth}}
\multicolumn{3}{c}{Variance ($\gamma$)}                                     \\ \hline
Source $\rightarrow$ Target & Original input          & Perturbed input          \\ \hline
CT $\rightarrow$ T2              & 0.0176 (0.0176)$\pm$0.0018 & 0.0197 (0.0195)$\pm$0.0024 \\
T2 $\rightarrow$ CT              & 0.0155 (0.0155)$\pm$0.0006 & 0.0199 (0.0195)$\pm$0.0031 \\
T1 $\rightarrow$ T2              & 0.0022 (0.0022)$\pm$0.0003 & 0.0031 (0.0028)$\pm$0.0010  \\
T2 $\rightarrow$ T1              & 0.0045 (0.0044)$\pm$0.0007 & 0.0104 (0.0115)$\pm$0.0048 \\
T1 $\rightarrow$ CT              & 0.0179 (0.0183)$\pm$0.0015 & 0.0185 (0.0185)$\pm$0.0027 \\
CT $\rightarrow$ T1              & 0.0567 (0.0573)$\pm$0.0034 & 0.0676 (0.0670)$\pm$0.0072  \\ \hline
\multicolumn{3}{c}{Confidence ($\delta$)}                                                \\ \hline
Source $\rightarrow$ Target & Original input          & Perturbed input          \\ \hline
CT $\rightarrow$ T2              & 2.2608 (2.2737)$\pm$0.0515 & 1.9398 (1.9529)$\pm$0.0939 \\
T2 $\rightarrow$ CT              & 2.7330 (2.7410)$\pm$0.0348 & 2.2738 (2.1684)$\pm$0.3819 \\
T1 $\rightarrow$ T2              & 2.0281 (2.0019)$\pm$0.0700 & 1.8230 (1.8364)$\pm$0.1736  \\
T2 $\rightarrow$ T1              & 2.2648 (2.2573)$\pm$0.0360 & 1.9289 (1.8636)$\pm$0.2300   \\
T1 $\rightarrow$ CT              & 2.7736 (2.7971)$\pm$0.0592 & 2.1120 (2.0829)$\pm$0.3806  \\
CT $\rightarrow$ T1              & 2.1806 (2.1818)$\pm$0.0517 & 1.7222 (1.6939)$\pm$0.1122
\end{tabular}
\label{tab:anomalies}
\end{table}

\subsection{Can we use the noise injections to improve the quality of a synthesized image?}
Our next experiment involved injecting noise into the training process. We investigated whether it would result in better image quality and a more robust representation. According to Table~\ref{tab:training_injections}, injecting a small amount of noise ($\alpha=0.5$) into half of the synthesized samples, during the training process, improved the final image quality. Nevertheless, it did not seem to translate to the end task: segmentation accuracy did not improve. We found out that introducing excess noise ($\alpha> 0.5$) or corrupting the majority of samples during training can cause confusion in the model, leading to deteriorated performance.

\begin{table}[t!]
\caption{Slightly corrupting the latent representation in GANs bottleneck leads to improved image quality. The lower the FID score, the better.}
\begin{tabular}{p{0.25\textwidth}p{0.17\textwidth}p{0.17\textwidth}p{0.17\textwidth}p{0.17\textwidth}}
\multicolumn{4}{l}{Image Quality [FID]}                                                             &            \\ \hline
Source $\rightarrow$ Target& \multicolumn{1}{l}{TarGAN}  & \multicolumn{1}{l}{\begin{tabular}[c]{@{}l@{}}TarGAN\\ ($\alpha = 0.5$)\end{tabular}} & \multicolumn{1}{l}{\begin{tabular}[c]{@{}l@{}}TarGAN\\ ($\alpha = 0.8$)\end{tabular}} & \multicolumn{1}{l}{\begin{tabular}[c]{@{}l@{}}TarGAN\\ ($\alpha = 1.0$)\end{tabular}} \\ \hline
T1 $\rightarrow$ CT              & 0.0649  & \textbf{0.0612} & 0.0616 & 0.0441                                                                                  \\
T2 $\rightarrow$ CT              & \textbf{0.1475}  & 0.1496 & 0.1704 & 0.1787                                                                                  \\
CT $\rightarrow$ T1              & 0.0651 & \textbf{0.0508} & 0.0584 & 0.0563                                                                                 \\
T2 $\rightarrow$ T1              & 0.1200  & \textbf{0.1135} & 0.1278 & 0.1450                                                                                  \\
CT $\rightarrow$ T2              & \textbf{0.0469} & 0.0471 & 0.0601 & 0.0647                                                                                 \\
T1 $\rightarrow$ T2              & 0.0604 & \textbf{0.0545} & 0.0583 & 0.0668                                                                                
\end{tabular}
\label{tab:training_injections}
\end{table} 

\subsection{Can we correlate our confidence score with the quality of downstream task, e.g., segmentation, on the synthesized image?}
To address this question, we train three U-Net~\cite{ronneberger2015unet} networks to perform liver segmentation on three imaging modalities, namely CT, T1 and T2, and then run the inference on both the same imaging modality and the transferred synthesized ones and report the results in Table~\ref{tab:segmentation}.  On the diagonal we present the scores for the original modality which range from 0.95 for CT to 0.82 for T1, which are slightly different from the ones reported in~\cite{chen2021targan} due to the fact of using a standard 2D U-Net and no enrichment technique~\cite{gupta2019enrichment}. Nevertheless, the segmentation results are acceptable for the CT to T1, CT to T2 and T1 to CT transferred images. However, the performance deteriorates for images where T2 scans are the source modality. This is reflected in the correlation scores as well (\emph{cf.} Table~\ref{tab:segmentation} and Table~\ref{tab:correlation}). There is a correlation around and higher than 0.5 for translations where the segmentor worked as well. This suggests that our method can be used most efficiently in cases where the images are generated well enough for the downstream task network to also perform well. If the generated images are of so low quality, that the segmentor fails completely (DICE < 0.5) the confidence value does not correlate with the DICE score.


\begin{table}[]
\caption{Segmentation results on original input images (diagonal) and images transferred with TarGAN. }
\begin{tabular}{p{0.10\textwidth}p{0.3\textwidth}p{0.3\textwidth}p{0.3\textwidth}}
\multicolumn{4}{l}{Segmentation quality {[}DICE{]}} \\  \hline
from\textbackslash{}to        & CT                      & T1                      & T2                      \\
CT                            & 0.9506 (0.9711)$\pm$0.1003 & 0.6806 (0.7315)$\pm$0.2225 & 0.7302 (0.7529)$\pm$0.1692 \\
T1                            & 0.6900 (0.8546)$\pm$0.3735   & 0.8276 (0.9578)$\pm$0.3134 & 0.5272 (0.6671)$\pm$0.4059 \\
T2                            & 0.4085 (0.5321)$\pm$0.3652 & 0.5088 (0.5302)$\pm$0.3662 & 0.8349 (0.9537)$\pm$0.2777
\end{tabular}
\label{tab:segmentation}
\end{table} 
\begin{table}[]
\caption{The FID scores and the absolute value of a correlation between the confidence score and DICE coefficient for our method and UP-GAN ~\cite{upadhyay2021uncerguidedi2i}.}
\begin{tabular}{p{0.2\textwidth}p{0.2\textwidth}p{0.2\textwidth}p{0.2\textwidth}p{0.2\textwidth}}
                                                         & \multicolumn{2}{l}{Noise injections} & \multicolumn{2}{l}{UP-GAN~\cite{upadhyay2021uncerguidedi2i}} \\ \hline
Source $\rightarrow$ Target & FID            & Correlation         & FID       & Correlation    \\ \hline
CT $\rightarrow$  T1                           & 0.0651         & 0.5423              & 0.2022    & 0.0029         \\
T1    $\rightarrow$ CT                           & 0.0649         & 0.5441              & 0.1619    & 0.2188         \\
CT    $\rightarrow$  T2                           & 0.0469         & 0.4946              & 0.1557    & 0.2540         \\
T2 $\rightarrow$  CT                           & 0.1475         & 0.2536              & 0.6540    & 0.0010         \\
T1   $\rightarrow$  T2                           & 0.0604         & 0.0546              & 0.1396    & 0.1021         \\
T2     $\rightarrow$  T1                           & 0.1200         & 0.3105              & 0.1656    & 0.0827        
\end{tabular}
\label{tab:correlation}
\end{table}

\subsection{How does the noise injection method compare to other uncertainty estimation techniques?}
We compare our method to the existing way of estimating aleatoric uncertainty, described in UP-GAN~\cite{upadhyay2021uncerguidedi2i}. The quality of generated images is measured with FID scores and the correlation between the DICE coefficient and the mean of the estimated aleatoric uncertainty values as defined
in \cite{upadhyay2021robustness}. It is not surprising that the FID scores are slightly lower than those of a TarGAN considering the absence of a shape-optimizing loss term. Furthermore, the aleatoric uncertainty does not correlate well with the DICE score, indicating that even though the aleatoric uncertainty might be useful in improving image quality as demonstrated in the paper, it does not translate directly into the downstream task of segmentation and cannot be used to indicate unsuitable samples. Among the differences between our method and \cite{upadhyay2021uncerguidedi2i}, we emphasize that ours only affects the inference stage and can be used with basically any architecture, while the UP-GAN involves significant differences in the architecture (extra outputs of the network) and the optimization process (extra loss terms requiring parameter tuning).

\section{Conclusion}
In this work, we investigated the hypothesis that a robust latent representation results in a higher quality of a generated image and higher performance on a downstream task. We showed that there are indicators that the quality of latent representation corresponds to the final quality of a generated image. If the downstream task network performs well, it is possible to correlate it with the latent representation's quality. Additionally, we discovered that small noise injections during the training phase lead to more robust representation and slightly higher image quality. However, this does not necessarily lead to better segmentation results. We compared the noise injections to the aleatoric uncertainty estimation method as proposed by \cite{upadhyay2021robustness}. Although our approach has a smaller impact on the image quality itself, it is more indicative of performance on downstream tasks. Our method is easier to incorporate as it does not require changes in the model's architecture or the optimization process.
The future work includes extending this method using adversarial attack techniques such as~\cite{ilyas2019adversarial,Zhang2021.12.23.21268289}, investigating how it influences different end tasks, for example classification, and validating the method in more real-life scenarios.


%
%
%
\bibliographystyle{splncs04}
\bibliography{refs}

\end{document}